\pdfoutput=1

\documentclass[11pt]{article}

\usepackage{acl}

\usepackage{times}
\usepackage{latexsym}
\usepackage{graphicx}
\usepackage{caption}
\usepackage{booktabs}
\usepackage{tabularx}
\usepackage{array} 
\usepackage{tabularray}
\usepackage{listings}
\usepackage{comment}
\usepackage{xcolor}
\usepackage{CJKutf8}
\usepackage{placeins}
\usepackage{todonotes}
\usepackage[utf8]{inputenc}
\usepackage{microtype}
\usepackage{inconsolata}
\usepackage{subfig}

\usepackage{enumitem}
\usepackage{amsmath}
\usepackage{amssymb}
\usepackage{booktabs}
\usepackage{multirow}
\usepackage{cleveref}
\usepackage{comment}

\usepackage{siunitx}
\usepackage{fvextra}
\DefineVerbatimEnvironment{promptblock}{Verbatim}{
  breaklines=true,
  breakanywhere=true,
  fontsize=\fontsize{7}{7},
  frame=single,
  framesep=1mm,
  breaksymbolleft={}
}

\newcommand\blfootnote[1]{%
  \begingroup
  \renewcommand\thefootnote{}\footnote{#1}%
  \addtocounter{footnote}{-1}%
  \endgroup
}

\usepackage[dvipsnames]{xcolor}

\newcommand{\pc}[1]{{\color{purple!70}pc: #1}}

\title{When Flores Bloomz Wrong: Cross-Direction Contamination\\ in Machine Translation Evaluation}

\author{
  \textbf{David Tan}\textsuperscript{1},
  \textbf{Pinzhen Chen}\textsuperscript{2,4},
  \textbf{Josef van Genabith}\textsuperscript{1,3},
  \textbf{Koel Dutta Chowdhury}\textsuperscript{1}
\\
  \textsuperscript{1}Saarland University, Saarland Informatics Campus\quad\textsuperscript{2}Queen's University Belfast\\
  \textsuperscript{3}German Research Center for Artificial Intelligence (DFKI)\quad\textsuperscript{4}University of Edinburgh
\\
 \texttt{data00003@stud.uni-saarland.de}}

\begin{document}
\maketitle
\begin{abstract}
Large language models (LLMs) can be benchmark-contaminated, resulting in inflated scores that mask memorization as generalization, and in multilingual settings, this memorization can even transfer to ``uncontaminated'' languages. Using the FLORES-200 translation benchmark as a diagnostic, we study two 7--8B instruction-tuned multilingual LLMs:  Bloomz, which was trained on FLORES, and Llama as an uncontaminated control.  
We confirm Bloomz's FLORES contamination and demonstrate that machine translation contamination can be cross-directional, artificially boosting performance in unseen translation directions due to target-side memorization. 
Further analysis shows that recall of memorized references often persists despite various source-side perturbation efforts like paraphrasing and named entity replacement.
However, replacing named entities leads to a consistent decrease in BLEU, suggesting an effective probing method for memorization in contaminated models.

\end{abstract}

\section{Introduction}

\blfootnote{Code at \href{https://github.com/Mr-Ao-25/cross-ling-contamination}{github.com/Mr-Ao-25/cross-ling-contamination}.}Large language models (LLMs) typically go through pre-training and fine-tuning stages to become applicable to downstream tasks, where data is a fundamental building block. Despite this, details about training data are often too vague to infer useful information, both for open- and closed-source models. This has negative implications for LLM research, one of which is the reliability of using public benchmarks to evaluate model performance. LLM training data is typically in the magnitude of billions to trillions of tokens \citep{gao2020pile, hoffmann2022training}. Consequently, public test sets may inadvertently be incorporated into training data \citep{sainz2024data} and artificially boost scores---a phenomenon referred to as \textit{data contamination} \citep{magar2022data}. The problem is even more pressing in multilingual evaluation, where contamination can cross-lingually transfer into languages not seen before \citep{yao-etal-2024-data}. 

This work uses machine translation as a diagnostic task to investigate cross-direction data contamination: the artificial boost of unseen language pair scores, resulting from training on language pairs with the same target. 
We use FLORES200 \citep{nllb2022}, a test suite that supports hundreds of languages with multiway parallelism, yet this wide coverage increases the likelihood of contamination. We explore whether its multiway parallel structure may result in cross-direction transfer of train-test leakage. 
We benchmark Bloomz-7B1 \citep{muennighoff2022crosslingual}, a multilingual LLM with FLORES documented as part of its fine-tuning data, and compare it to Llama-3.1-8B-Instruct \citep{grattafiori2024llama3herdmodels} 
which is reportedly clean. Compared to \citet{kocyigit2025overestimation}'s focus on the contribution of source or target-side text during training, we probe cross-direction contamination in machine translation at inference, specifically whether altered source inputs can still trigger memorized targets. 
We analyze patterns of contamination through controlled studies and design perturbation-based tests to probe how contamination manifests. We present three findings:
\begin{enumerate}[itemsep=0ex,topsep=0ex,parsep=0ex]
    \item We show that contamination happens cross-directionally for machine translation, where contaminated LLMs can perform moderately well in unseen translation directions, mainly due to target-side language memorization. 
    \item While it is assumed that train-test source input similarity leads to patterns of contamination, we demonstrate that a not-so-similar input may still lead to a model calling memorized text. 
    \item We discover that replacing named entities consistently decreases BLEU, suggesting an effective method for probing memorization.
\end{enumerate}

\begin{figure*}[t]
  \centering\small
  \begin{tabular}{m{0.43\linewidth}m{0.43\linewidth}}
    \centering\includegraphics[width=\linewidth]{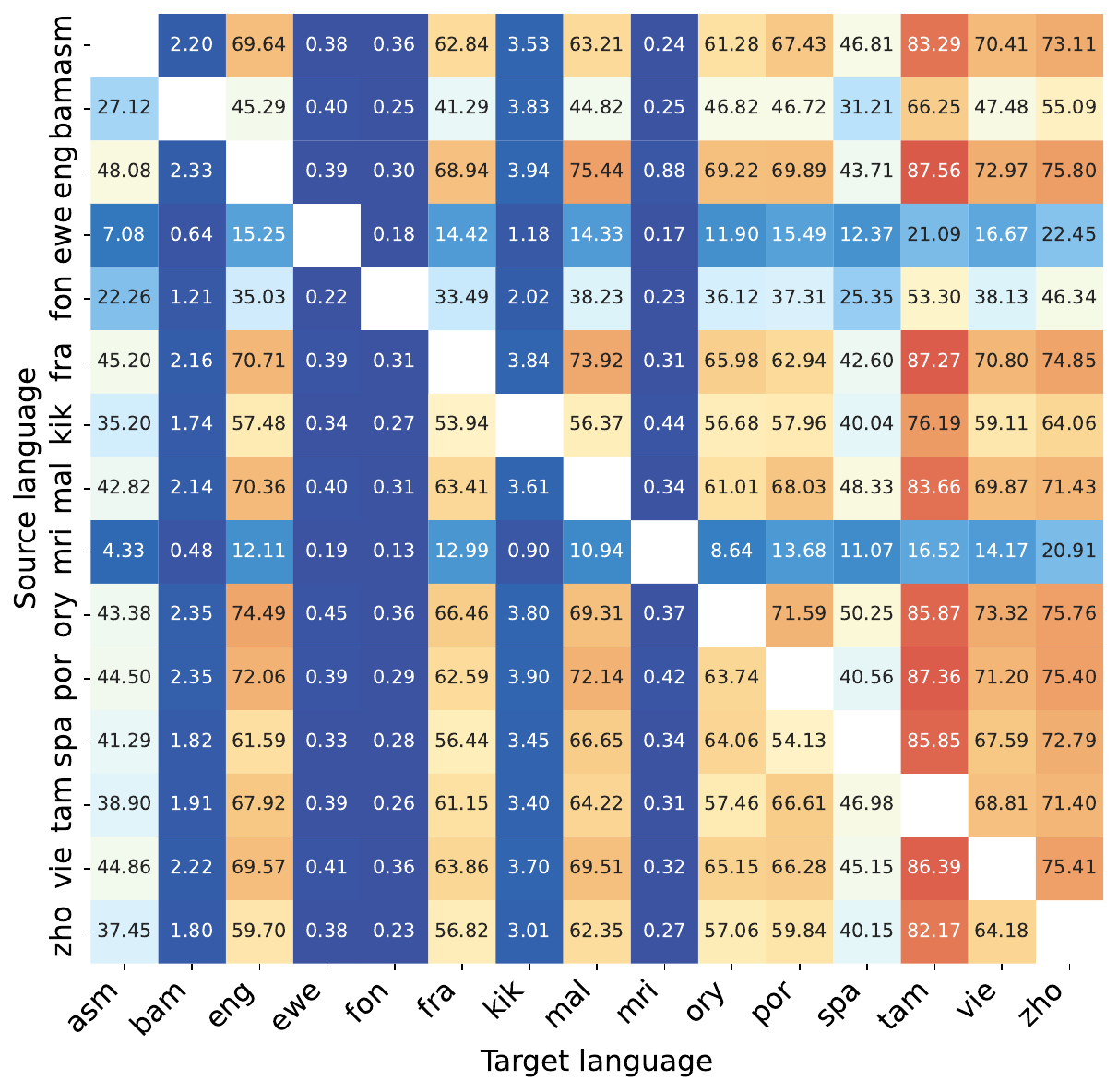} 
    \centering\includegraphics[width=\linewidth]{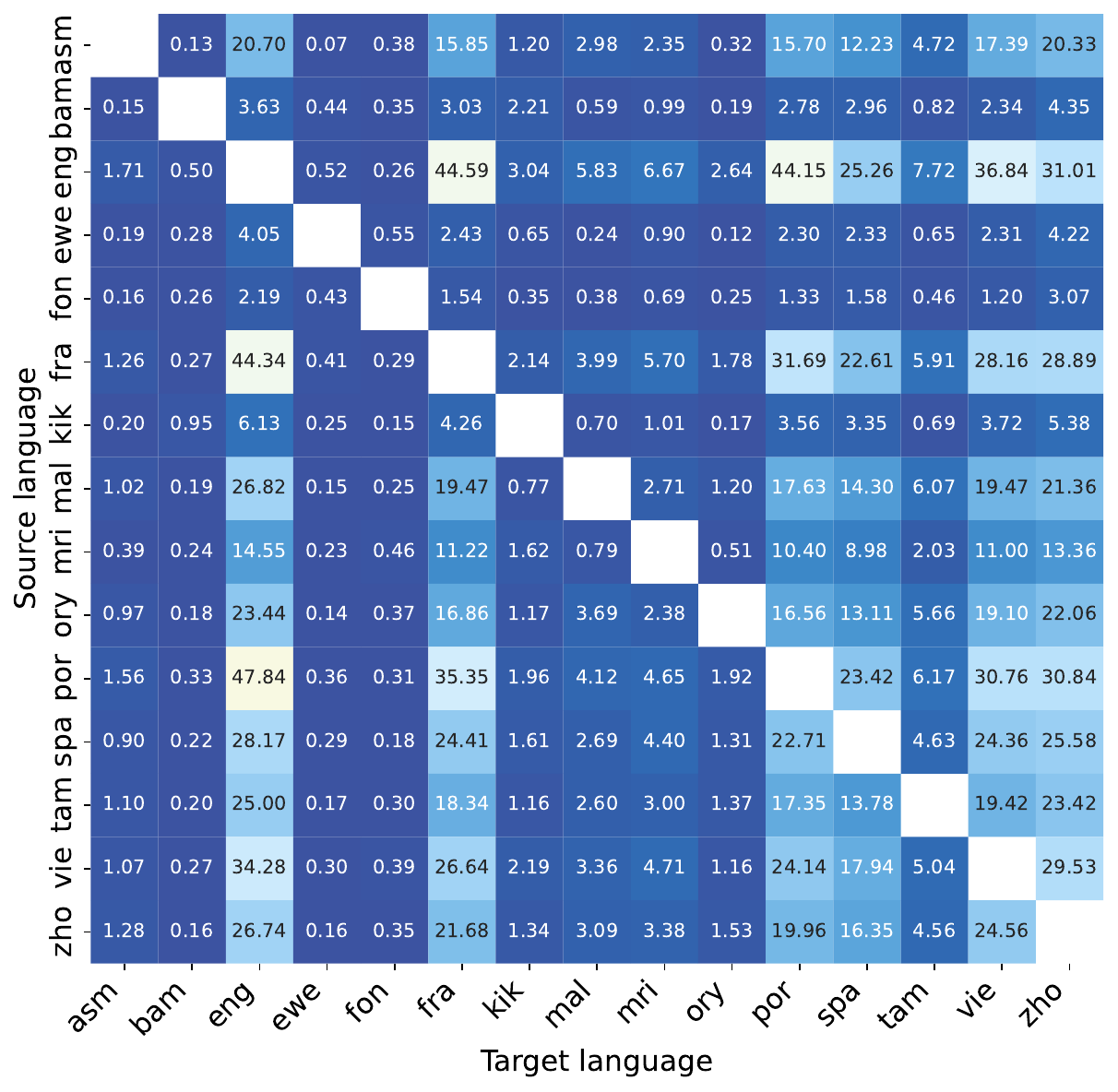} &
    \centering\includegraphics[width=\linewidth]{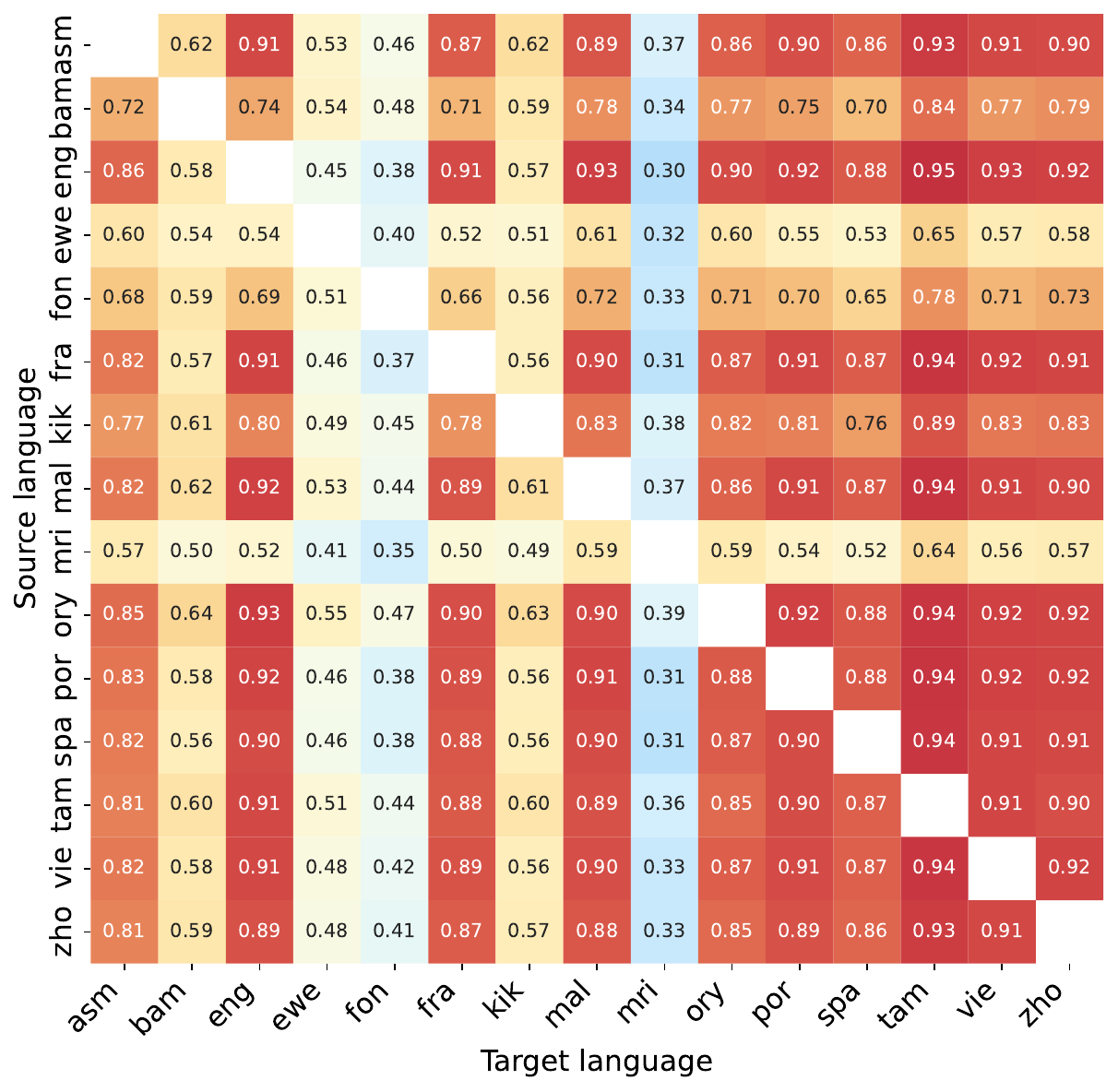}
    \centering\includegraphics[width=\linewidth]{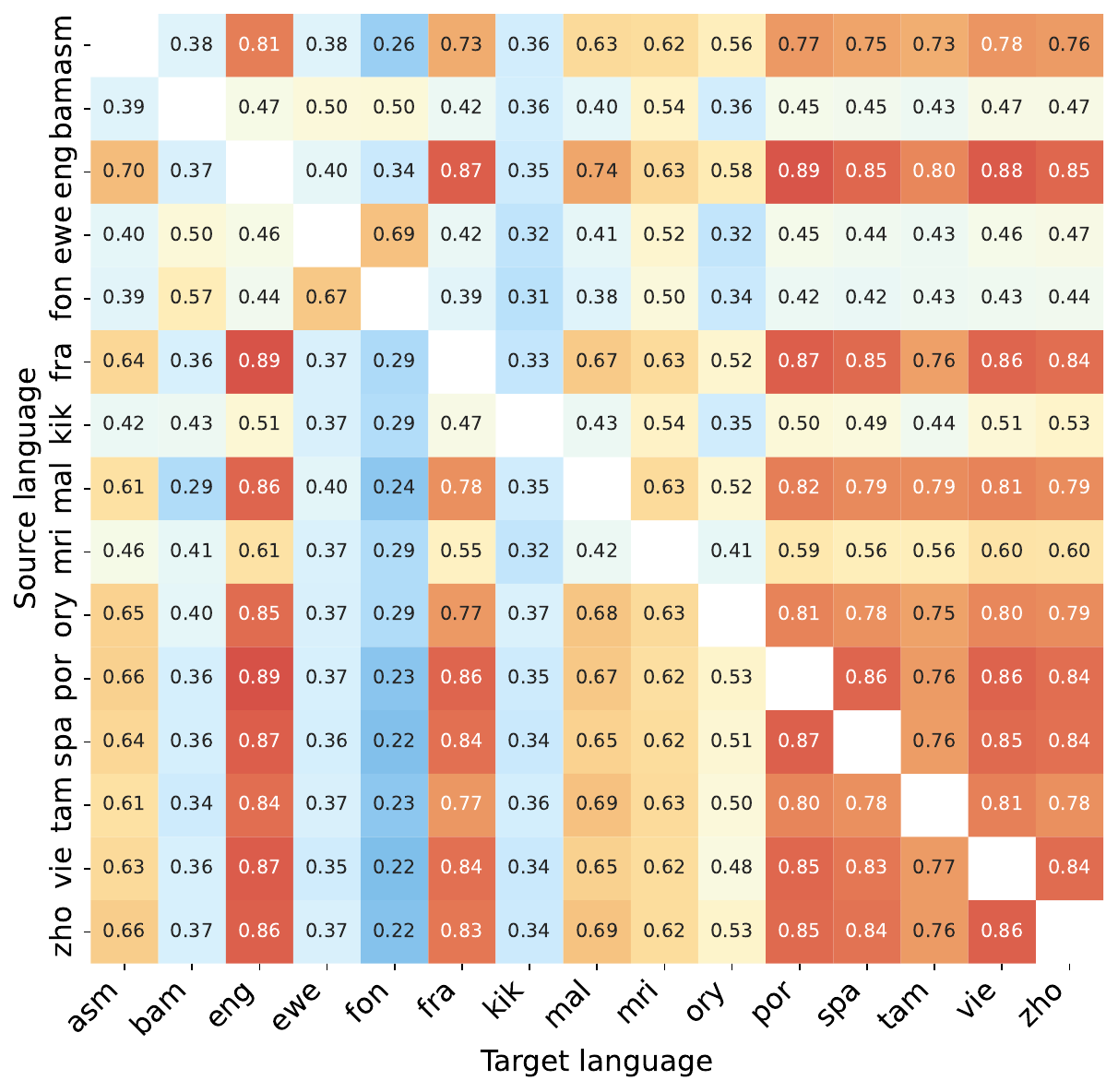}
  \end{tabular}     
  \caption{BLEU for Bloomz (upper left), Llama (lower left) and COMET for Bloomz (upper right), Llama (lower right). All plots have the same sources, targets, and color bar with a blue-red gradient (low-to-high).}
  \label{fig:heatmaps}
\end{figure*}

\section{Experimental Setup}
\paragraph{Test set and Models} We use FLORES200 \citep[][or \textit{FLORES}]{nllb2022} as our test set for its wide language coverage and multi-way parallelism. We use the dev set of 997 samples and evaluate the model’s initial generation for each sample. We select 15 languages spanning different resource levels—high-resourced: English (\texttt{eng}), Simplified Chinese (\texttt{zho}), Spanish (\texttt{spa}), Portuguese (\texttt{por}), French (\texttt{fra}), and Vietnamese (\texttt{vie}); medium-resourced: Malayalam (\texttt{mal}), Tamil (\texttt{tam}), Assamese (\texttt{asm}), and Odia (\texttt{ory}); low-resourced: Bambara (\texttt{bam}), Fon (\texttt{fon}), Ewe (\texttt{ewe}), Kikuyu (\texttt{kik}), and Maori (\texttt{mri}). Notably, Bloomz has not been trained on \texttt{ewe} and \texttt{mri}.



We investigate various aspects of contamination using \texttt{bloomz-7b1} \cite[][\textit{Bloomz}]{muennighoff2022crosslingual} and \texttt{Llama-3.1-8B-Instruct} \citep[][\textit{Llama}]{grattafiori2024llama3herdmodels}. 
Both models are instruction-tuned, similarly sized (7–8B parameters), and designed for multilingual text generation, making them broadly comparable in their cross-lingual capabilities. While Bloomz was explicitly trained on FLORES with unspecified translation directions, there is no reported use of FLORES in the training of Llama.
For entity replacement experiments, we use \texttt{aya-expanse-8b} \cite[][\textit{Aya}]{dang2024ayaexpansecombiningresearch}.

\section{Contamination in Machine Translation}

\subsection{Is there contamination?}
\label{sec:3.1}
We first investigate whether Bloomz and Llama have memorized FLORES. We run an evaluation on FLORES for every source-target pair formed by our 15 chosen languages. 
We use BLEU implemented in sacrebleu \citep{post-2018-call,papineni-etal-2002-bleu} as a measure of surface form overlap w.r.t. the reference, rather than translation quality, and COMET-22-DA \citep{rei-etal-2022-comet} for semantic similarity. As our diagnostic, we take that an unreasonably high BLEU coupled with high COMET (e.g. 80 BLEU, 0.9 COMET) would imply contamination, while a reasonable BLEU with a high COMET score (e.g. 40 BLEU, 0.9 COMET) would indicate a good performance without contamination (or with less likelihood of it). Low BLEU and COMET indicate divergence in both form and meaning; moderate BLEU with high COMET reflects semantic similarity despite surface variation.

\Cref{fig:heatmaps} presents BLEU and COMET heatmaps for Bloomz and Llama. 
For Bloomz, we see oddly high BLEU (40--90) in \texttt{xxx}$\to$\{\texttt{eng}, \texttt{fra}, \texttt{mal}, \texttt{ory}, \texttt{por}, \texttt{tam}, \texttt{vie}, \texttt{zho}\}, hinting at contamination; medium BLEU scores ($\leq$ 50) in \texttt{xxx}$\to$\{\texttt{asm}, \texttt{spa}\}; and very low scores ($\approx$ 0) in \texttt{xxx}$\to$\{\texttt{bam}, \texttt{ewe}, \texttt{fon}, \texttt{kik}, \texttt{mri}\}. COMET scores generally follow trends seen in BLEU; language pairs with medium to high BLEU achieve $>$ 0.8 COMET; low BLEU language pairs achieve $\leq 0.5$ -- 0.65 COMET.

In contrast, Llama shows little evidence of contamination: reasonably high BLEU scores (30--40) only appear in a few high-resource languages or directions, and COMET scores are only high ($\geq$ 0.8) when BLEU scores are $>$15. 
For Bloomz, high COMET ($\geq$ 0.8) results from $\geq$ 35 BLEU.


\paragraph{Memorization not generalization} To rule out the possibility that Bloomz simply has superb performance in those directions with high BLEU, we test the relatively low-resource \texttt{eng}$\to$\{\texttt{mal}, \texttt{ory}, \texttt{tam}\} available in PMIndia \cite{2020arXiv200109907H} and Mann-ki-Baat \cite{jain2024bhasaanuvaad}. 
\Cref{pmindia_mann-ki-baat} \Cref{tab:pmindia--man-ki-baat} shows near-zero BLEU scores for both datasets, indicating that Bloomz cannot translate those directions, adding further evidence of FLORES contamination.

\paragraph{Disentangling source and target}
We study the pattern of contamination when a language is placed on the source or target side. \Cref{fig:heatmaps} shows that when placed on the target side, a handful of languages have a ``clean'' column: \texttt{bam}, \texttt{ewe}, \texttt{fon}, \texttt{kik}, and \texttt{mri}, with BLEU $\approx$ 0. 
In contrast, when placed on the source side, a few languages that did not exhibit contamination signals as the target (\texttt{bam}, \texttt{fon}, \texttt{kik}) now have medium-range BLEU scores across their rows. Overall, no single language as a source can escape contamination: even \texttt{ewe} and \texttt{mri} that are not supported by Bloomz achieve $>$10 BLEU into a few target languages, higher than Llama. This suggests that contamination manifests itself asymmetrically in terms of whether a language is at the source or target side: inflated scores are due to the memorization of the target language text.


\subsection{Is the recall due to the exact source input?}

We investigate whether Bloomz' recall of the memorized target is due to seeing the \textit{exact} source string paired with the target. 
For all \texttt{xxx}$\to$\texttt{yyy} translation directions, we utilize a back-translated source, \texttt{xxx}\textsubscript{\texttt{zzz}}, originating from a third language \texttt{zzz}; this is essentially done by Llama in \texttt{zzz}$\to$\texttt{xxx} directions in \Cref{sec:3.1}.
FLORES' multiway parallelism ensures that the back-translated source remains parallel to the original target reference, while staying different from the original source.

We try three originating languages \texttt{zzz}=\{\texttt{asm}, \texttt{bam}, \texttt{por}\}, for which Llama had distinct translation performance (into any \texttt{xxx}): \texttt{xxx}\textsubscript{\texttt{bam}} are near-zero BLEU, \texttt{xxx}\textsubscript{\texttt{asm}} are in the 10--20 range, and \texttt{xxx}\textsubscript{\texttt{por}} are in the 20--50 range.
These scores are shown in Llama's heatmap in \Cref{fig:heatmaps}. 
Specifically, we experiment with all chosen languages, except \{\texttt{por}, \texttt{bam}, \texttt{asm}\}, as source \texttt{xxx}. 
The target language is fixed to \texttt{tam}, which exhibited the highest degree of target-side memorization. This forms three test directions: \{\texttt{xxx}\textsubscript{\texttt{por}}, \texttt{xxx}\textsubscript{\texttt{asm}}, \texttt{xxx}\textsubscript{\texttt{bam}}\}$\to$\texttt{tam}. 

We also utilize a paraphrased source generated by Llama denoted as \texttt{xxx}\textsubscript{pp} to compare against the Llama back-translated \texttt{xxx}\textsubscript{zzz}. We test high-resource source languages into \texttt{tam}: \texttt{xxx}\textsubscript{pp} $\to$ \texttt{tam} and compare it against \texttt{xxx}\textsubscript{por} $\to$ \texttt{tam}. We also assess the similarity between the altered sources and the original source denoted as \texttt{xxx}\textsubscript{og}: \{\texttt{xxx}\textsubscript{por}, \texttt{xxx}\textsubscript{pp}\} $\to$ \texttt{xxx}\textsubscript{og}.

\begin{figure}[t]
\centering\small
  \includegraphics[width=0.6\linewidth]{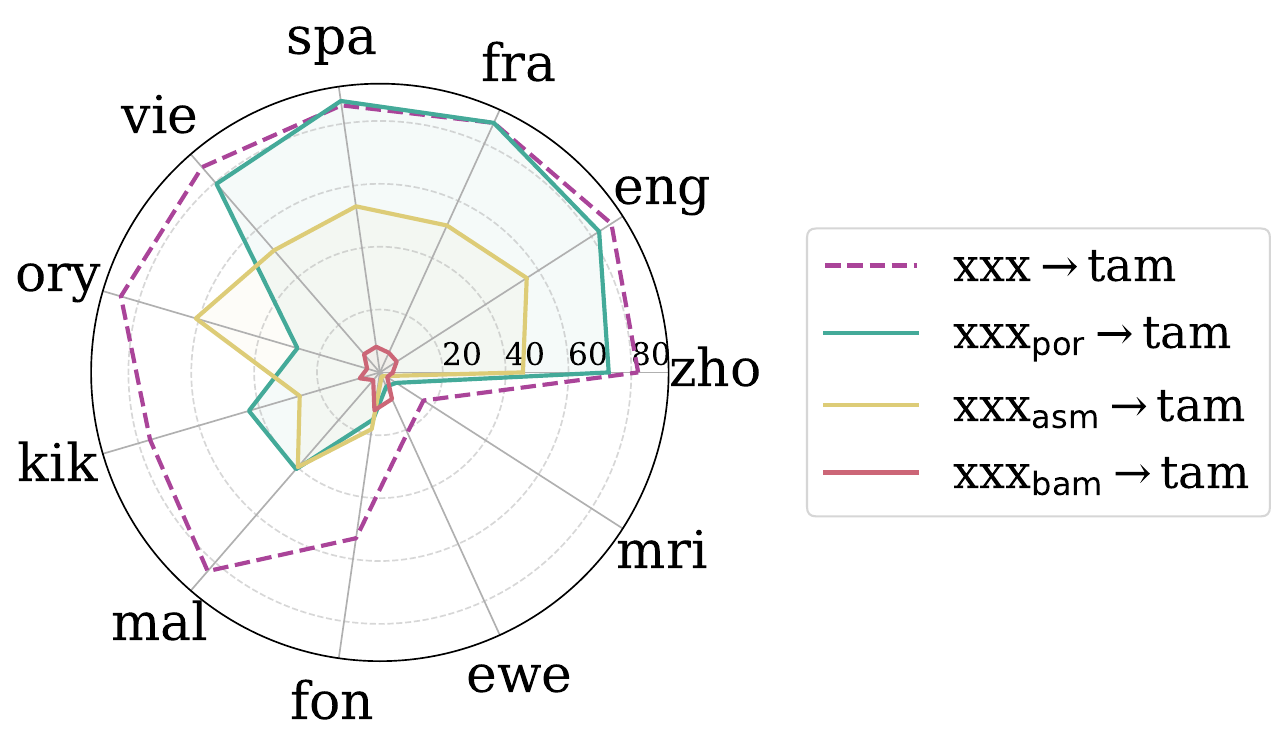}
  \caption{Bloomz's BLEU for \{\texttt{xxx}\textsubscript{\texttt{por}}, \texttt{xxx}\textsubscript{\texttt{bam}}, \texttt{xxx}\textsubscript{\texttt{asm}}\}$\to$\texttt{tam}. Llama back-translated \{\texttt{por}, \texttt{bam}, \texttt{asm}\} into \texttt{xxx}\textsubscript{\{\texttt{por}, \texttt{bam}, \texttt{asm}\}}.}
  \label{fig:pivot translation}
\end{figure}
\begin{figure}
\centering\small
  \includegraphics[width=1\linewidth]{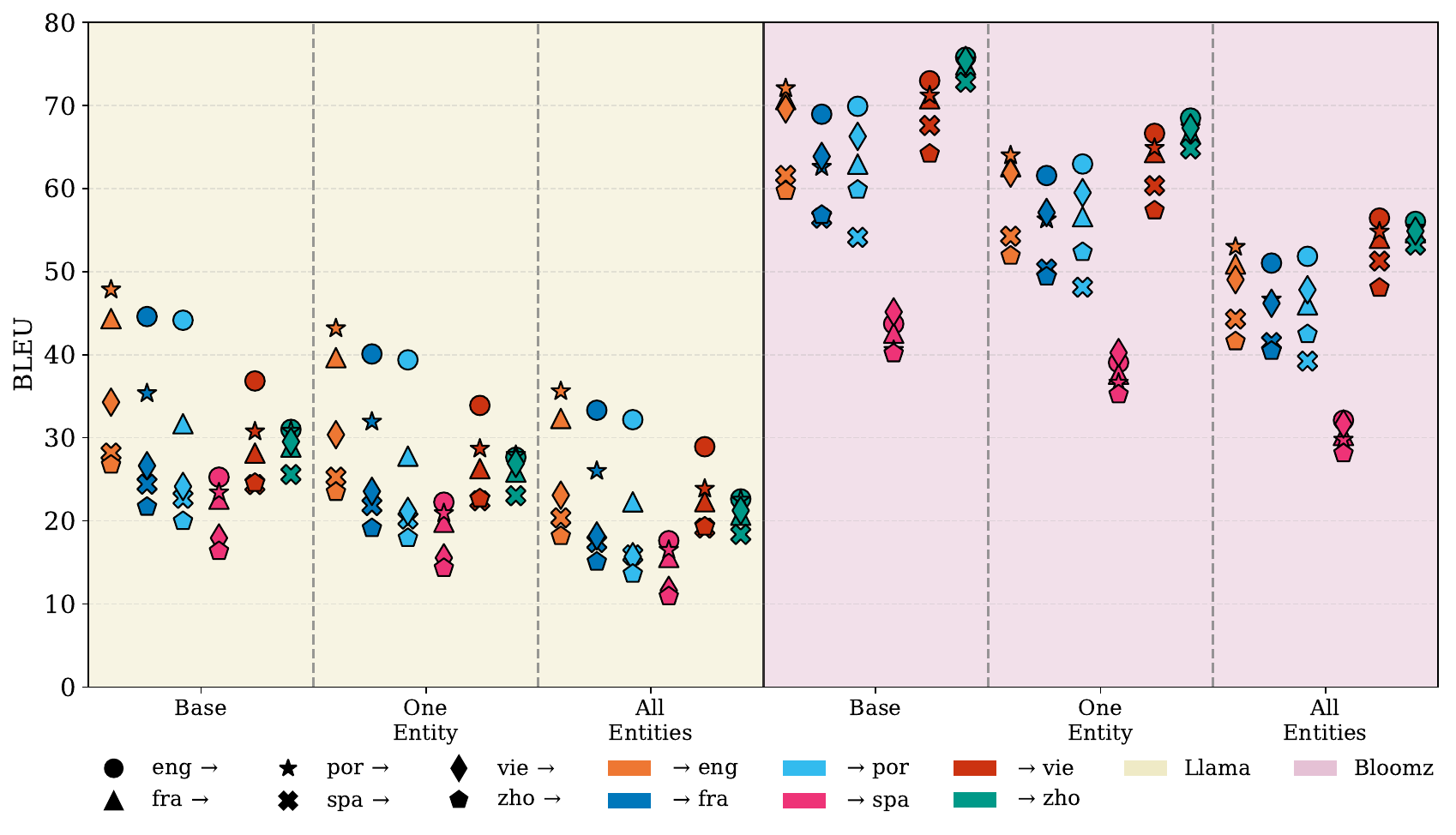}
  \caption{BLEU scores for Llama (left) and Bloomz (right) across language pairs under different entity replacement settings (Base, One Entity, All Entities).}
  \label{fig:perturbation}
\end{figure}



\paragraph{Results} \Cref{fig:pivot translation} presents Bloomz' performance in translating back-translated source. 
Generally, the BLEU scores of  \texttt{xxx}\textsubscript{src}$\to$\texttt{tam} align with the performance of Llama when producing the back-translations (\texttt{xxx}\textsubscript{por}, \texttt{xxx}\textsubscript{bam}, \texttt{xxx}\textsubscript{asm}). 
When Llama's back-translation is in the 20--50 BLEU range, Bloomz' translation has a high $>$60 BLEU, as exhibited in \{\texttt{vie}, \texttt{spa}, \texttt{fra}, \texttt{eng}, \texttt{zho}\}\textsubscript{\texttt{por}}$\to$\texttt{tam}. When the string overlap between the back-translated and original sources is lower (10--20 BLEU), Bloomz gets a moderate 45--50 BLEU as exhibited in \{\texttt{vie}, \texttt{spa}, \texttt{fra}, \texttt{eng}, \texttt{zho}\}\textsubscript{\texttt{asm}}$\to$\texttt{tam}. 

Unexpected results are observed in a few back-translated sources with close to 0 
BLEU (i.e.\ \{\texttt{ory}, \texttt{kik}, \texttt{mal}\}\textsubscript{\{\texttt{por}, \texttt{asm}\}}). 
Given these back-translated sources, Bloomz still yield 20--60 BLEU, indicating moderate to high recall of memorized targets.
In other cases, where the back-translated source has poor BLEU w.r.t. the original source, memorized target recall is minimal, as shown in other low-resource sources from \texttt{por} or \texttt{asm}, as well as in all \texttt{xxx}\textsubscript{bam}$\to$\texttt{tam}. 

\Cref{tab:pp_vs_backtrans} presents the paraphrased scores compared against the back-translated scores. Despite the paraphrased sources having higher surface form overlap with the original, the back-translated sources score 20--30 BLEU higher for all pairs. Our results suggest that the exact source is the culprit, but even a not-so-similar back-translated or paraphrased input does not guarantee a low recall, indicating that some aspect of the source functions as a "trigger" for recall.

\subsection{What features contribute to the recall?}

We investigate the curious case above, where back-translated sources still lead to recall of memorized target. 
First, we manually inspect Llama's back-translations (refer to \Cref{tbl: output examples} in \cref{Appendix llama backtrans}) and observe that even when they diverge from the original sources significantly, named entities are still retained.  To investigate whether named entities contribute to recall, we create new entities to replace those in the original using six high resource languages (\texttt{eng}, \texttt{fra}, \texttt{por}, \texttt{spa}, \texttt{vie}, \texttt{zho}) and observe how the recall of memorized targets varies, using Bloomz' and Llama's BLEU scores.

\begin{figure*}[t]
  \centering\small
  \begin{tabular}{m{0.4\linewidth}m{0.4\linewidth}}
    \centering\includegraphics[width=\linewidth]{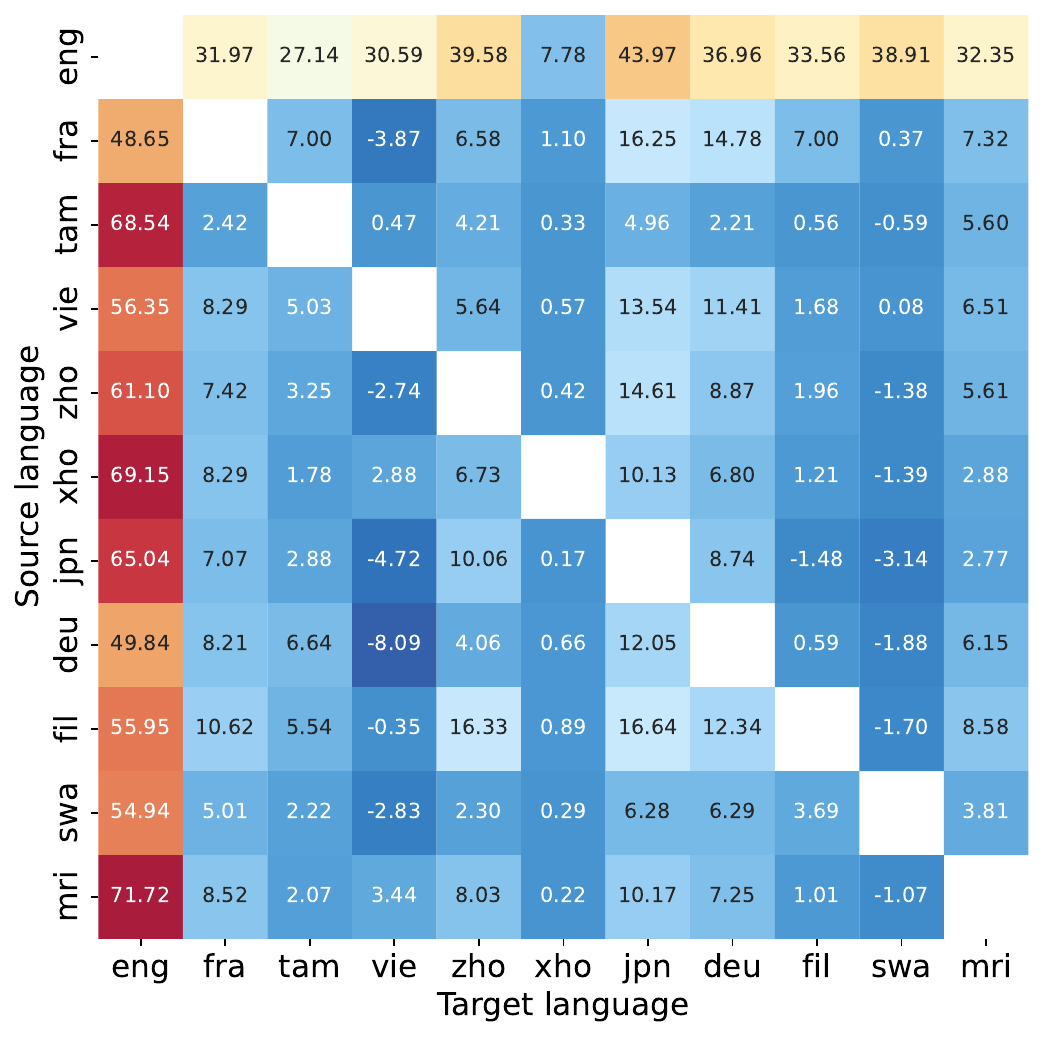} &
    \centering\includegraphics[width=\linewidth]{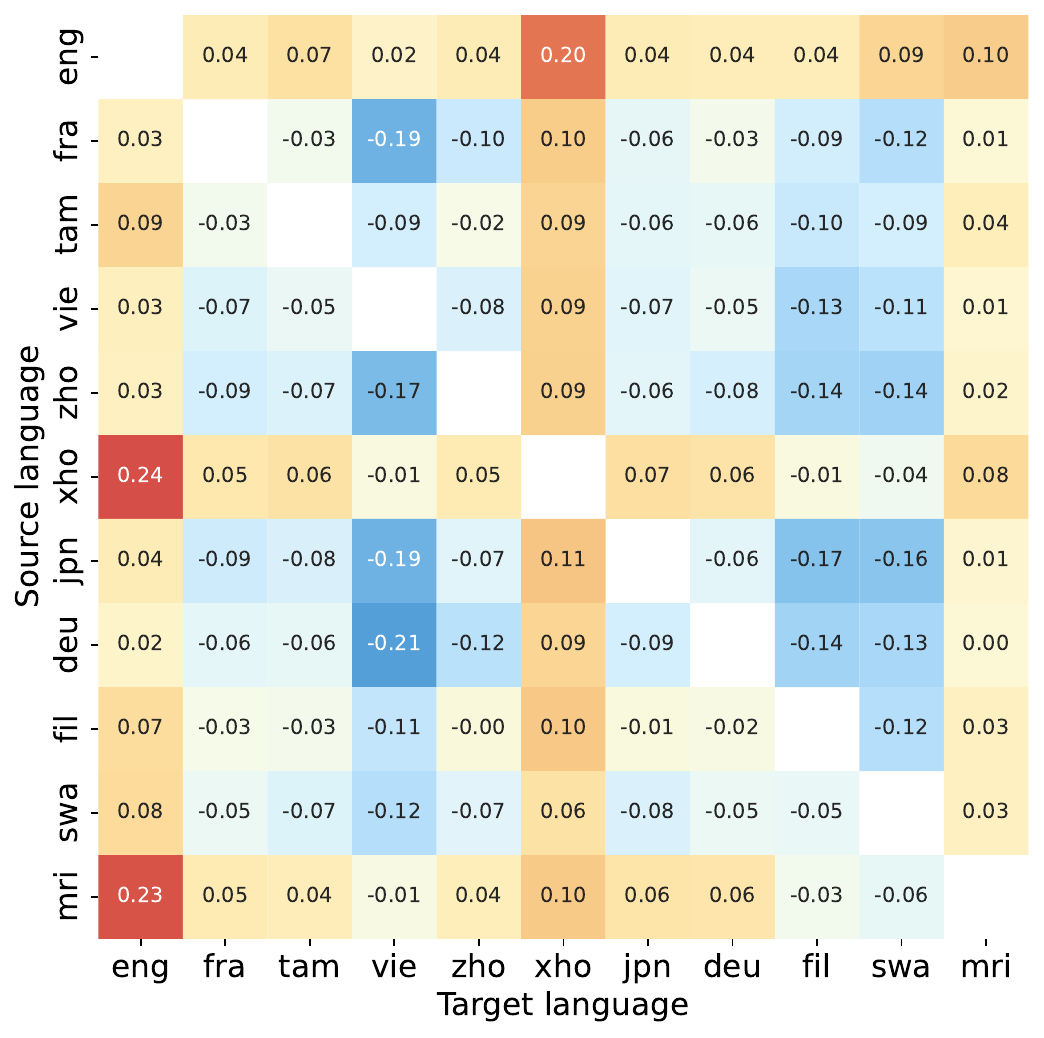}
    \end{tabular}     
  \caption{The BLEU (left) and COMET (right) difference between the finetuned and base Llama model. Positive/negative indicates increases/decreases in the fine-tuned model compared to the base model.}
  \label{fig:ft_heatmaps}
\end{figure*}
To ensure that the same named entities are identified and labeled for every language, we use spaCy \citep{Honnibal_spaCy_Industrial-strength_Natural_2020} to first identify named entities for every sentence and manually edit any missing or incorrectly labeled entities. Given the English source and its corresponding list of entities, we use Aya to create new entities to replace the original and translate the new entities into every high resource language. For entity types, descriptions, and examples of entity-replaced sentences, refer to \Cref{tbl: entity replaced examples} and \Cref{tab:appendix:entity_descriptions}.
In total, FLORES contains around 640 sentences with entities for our chosen languages.

We experiment with two replacement settings, one-entity and all-entities, and compare against the original for each language pair. 
For the one-entity setting, sentences with multiple entities result in multiple variants, from which one is chosen randomly. The all-entities setting replaces every entity in a sentence. We then score BLEU for each language pair against the original reference to measure recall of memorized text.
\begin{table}[t]
\small\centering
\setlength{\tabcolsep}{1.2ex}
\begin{tabular}{lSSSSS}
\toprule
 & \texttt{eng} & \texttt{fra} & \texttt{spa} & \texttt{vie} & \texttt{zho}      \\
\midrule
\texttt{xxx}\textsubscript{por} $\to$ \texttt{tam}   &  82.92 & 87.28 & 87.24 & 79.38 & 72.81  \\ 
\texttt{xxx}\textsubscript{pp} $\to$ \texttt{tam}      & 64.60 & 66.55 & 61.15 & 62.18 & 53.67 \\
\midrule
\texttt{xxx}\textsubscript{por} $\to$ \texttt{xxx}\textsubscript{og}  & 47.83 & 35.35 & 23.41 & 30.75 & 30.84 \\
\texttt{xxx}\textsubscript{pp} $\to$ \texttt{xxx}\textsubscript{og} & 53.03 & 60.92 & 54.38 & 62.60 & 61.11 \\
\bottomrule
\end{tabular}
\caption{\{\texttt{xxx}\textsubscript{por}, \texttt{xxx}\textsubscript{pp}\} $\to$ \texttt{tam} evaluated by Bloomz and the perturbed sources compared against the original source: \{\texttt{xxx}\textsubscript{por}, \texttt{xxx}\textsubscript{pp}\} $\to$ \texttt{xxx}\textsubscript{og}}
\label{tab:pp_vs_backtrans}
\end{table}
\paragraph{Results} {\Cref{fig:perturbation} shows the changes in BLEU for both models when entity replacement is applied. Generally, results indicate that replacing entities in source texts are associated with a reduced level of recall. Bloomz results in a reduction of 5-10 BLEU for single entity replacement and 10-20 BLEU for all entity replacement. Llama experiences an average decrease of about 5 BLEU for the one-entity setting and about 10 for the all-entities setting, with certain pairs decreasing by up to 15 BLEU. 
For each target language, the relative ordering of scores across language pairs is largely preserved, confirming that target-side memorization remains the primary factor. However, the consistent BLEU drop makes entity replacement a potentially effective method for probing memorization in parallel datasets like FLORES.

\section{Is contamination occurring cross-directionally?}

To investigate whether the multilingual nature of FLORES inadvertently causes cross-direction contamination during training, we perform fine-tuning experiments with FLORES \texttt{eng} $\leftrightarrow$ \texttt{xxx} data in 11 languages 
\{\texttt{eng}, \texttt{fra}, \texttt{tam}, \texttt{vie}, \texttt{mri}, \texttt{zho}, Xhosa (\texttt{xho}), Japanese (\texttt{jpn}), German (\texttt{deu}), Tagalog (\texttt{fil}), Swahili (\texttt{swa})\} 
using Axolotl.\footnote{\href{https://github.com/axolotl-ai-cloud/axolotl}{github.com/axolotl-ai-cloud/axolotl}} For the given languages, \texttt{xxx}, \texttt{yyy}, and \texttt{zzz}, we define cross-direction contamination as the artificial boost in scores of unseen translation directions (\texttt{zzz} $\to$ \texttt{yyy}) due to the memorization of \texttt{yyy} in seen translation directions (\texttt{xxx} $\to$ \texttt{yyy}). Although exact translation directions used for model training are rarely specified, it is reasonable to assume English is used as both source and target.

We fine-tune Llama until it severely memorizes the \texttt{eng} $\leftrightarrow$ \texttt{xxx} data to mirror the level of memorization observed in Bloomz (refer to \Cref{tab:appendix:fine-tune parameters} for fine-tuning parameters). 
Using the fine-tuned model, we translate every source-target pair for the 11 languages and evaluate using BLEU and COMET.

\paragraph{Results}
\Cref{fig:ft_heatmaps} presents the BLEU and COMET differences between the fine-tuned and base models (refer to \Cref{fig:llama_ft_heatmaps_appendix} for full scores).
In general, BLEU scores for most unseen language pairs increased up to 16 points, the exceptions being \texttt{xxx} $\to$ \{\texttt{vie}, \texttt{swa}\}. As for \texttt{eng} $\leftrightarrow$ \texttt{xxx}, the gains in \texttt{xxx} $\to$ \texttt{eng} overshadow the gains in \texttt{eng} $\to$ \texttt{xxx} reinforcing target side memorization.

Concerning source and target memorization, we observe trends mirroring Bloomz. Although \texttt{xho} yielded low BLEU scores as a target, its effectiveness as a source was similar to other languages.
Likewise, \texttt{xxx} $\to$ \{\texttt{vie}, \texttt{swa}\} showed decreases with most sources, but performed
similarly to other languages when used as sources, reinforcing that contamination is due to target-side memorization.

Conversely, the trends in COMET contradict the overall increases in BLEU. Outside of \texttt{xxx} $\to$ \{\texttt{mri}, \texttt{xho}\}, COMET scores decreased for all pairs. We conjecture  this is due to the extreme memorization of some samples, leading to increased overall BLEU scores at the cost of the non-memorized samples. Regarding the exceptions \{\texttt{mri}, \texttt{xho}\}, we surmise that they could not deteriorate further, as the BLEU and COMET in the base model already indicated little similarity with the reference.

\section{Conclusion}
We verified that Bloomz is FLORES-contaminated, using Llama as an uncontaminated control, and demonstrate through fine-tuning Llama with FLORES \texttt{eng}$\leftrightarrow$ \texttt{xxx} pairs that contamination is cross-directional, driven by target-side memorization.
We further show that back-translated and paraphrased sources do not necessarily reduce the recall of memorized references. We found that replacing named entities leads to a consistent decrease in BLEU, suggesting an effective method for probing memorization in potentially contaminated models. We recommend that practitioners verify contamination across translation directions when evaluating on multiway parallel benchmarks.

\newpage
\section*{Acknowledgment}
We thank Barry Haddow for his comment on paraphrasing the Flores input. The work received funding from Deutsche Forschungsgemeinschaft (DFG, German Research Foundation) – SFB 1102 Information Density and Linguistic Encoding.

\section{Limitations}
Although we gave Aya the source, source entities, and target during entity generation, it is possible that the new entities have different inflection compared to the original entities, changing the sentence more than intended. The drop in BLEU for entity replacement may differ depending on the number of sentences that contain entities. Also, our work yielded empirical conclusions but did not investigate the internal mechanisms of cross-direction contamination or the activation of memorized text. Our experiments are limited to 7-8B parameter models; results may vary for different model sizes. Additionally, our fine-tuning experiments simulate contamination but may not fully replicate contamination that occurs during pre-training.


\section{Ethical Considerations}
This study sought to uncover various forms of contamination in machine translation tasks. Our work poses no risk to readers, practitioners, or the wider community. All model and data artifacts were used in full compliance with their respective licenses.

\bibliography{custom}

\newpage

\appendix

\section{Related Work}

Recent studies highlighted the severe implications of data contamination and called for increased scrutiny of this issue \citep{pan2020privacy, zhou2023don, jacovi2023stop, dodge2021documenting}. 
LLMs exhibit substantial memorization capabilities \citep{elangovan2021memorization, hartmann2023sok, carlini2023quantifying}, raising concerns that benchmark gains may reflect leakage rather than generalization. To address this, prior work proposed both post hoc detection \citep{shi2023detecting, oren2023proving} and controlled experiments where models are deliberately trained with test data to measure their effects \citep{jiang2024investigating, yang2023rethinking, magar2022data}. 

In machine translation (MT), memorization and contamination manifest differently. \citet{raunak2021curious} documented that hallucinations often arise from memorized artifacts. \citet{raunak2022finding} showed the phenomenon of extractive memorization, where models reproduce target segments after partial source exposure. \citet{raunak2022salted} linked memorizing rare patterns to long-tail translation errors. 
\citet{guerreiro2022looking} provided a systematic analysis of hallucinations, showing how memorization failures scale across multilingual settings. \citet{chowdhury2020understanding, dutta-chowdhury-etal-2021-tracing} demonstrated that translation artifacts manifest as detectable patterns in embedding spaces. More recently, \citet{kocyigit2025overestimation} conducted controlled pre-training and found that accidental inclusion of evaluation examples in pre-training corpora critically undermines validity. 
Taken together, prior research has focused on contamination and memorization in monolingual or bilingual MT. To our knowledge, our work is the first to systematically examine cross-lingual contamination in MT, demonstrating that memorization in the target language can artificially boost performance in translating many source languages.

\FloatBarrier
    
\begin{table}[t]
\section{PMIndia and Maan-ki-Baat BLEU Scores}
\label{pmindia_mann-ki-baat}
\small\centering
\begin{tabular}{lSSS}
\toprule
Test set      & {\texttt{eng}$\to$\texttt{mal}}       & {\texttt{eng}$\to$\texttt{ory}}      & {\texttt{eng}$\to$\texttt{tam}}       \\
\midrule
FLORES    &  75.44 & 69.21 &  87.55  \\ 
PMIndia      & 0.12 &  1.85 &  0.11 \\
Mann-ki-Baat  & 0.26 &  1.01 & 0.82 \\
\bottomrule
\end{tabular}
\caption{Bloomz's BLEU for \texttt{eng}$\to$\{\texttt{mal}, \texttt{ory}, \texttt{tam}\}.}
\label{tab:pmindia--man-ki-baat}
\end{table}

\begin{table}[t]
\section{Fine-tuning Parameters}
\label{Appendix Fine-tuning}
\vspace{2ex}
\centering
\begin{promptblock}
base_model: meta-llama/Meta-Llama-3.1-8B-Instruct
model_config:
    attention_dropout: 0.1
sequence_len: 512
sample_packing: true
pad_to_sequence_len: true
num_epochs: 3
micro_batch_size: 4
gradient_accumulation_steps: 8  
learning_rate: 5e-6
weight_decay: 0.05
warmup_ratio: 0.02
lr_scheduler: cosine
max_grad_norm: 0.25
optim: adamw_torch
deepspeed: deepspeed_configs/zero3_bf16.json
torch_distributed_type: DEEPSPEED
\end{promptblock}
\vspace{-2ex}
\caption{Fine-tuning Parameters}
\label{tab:appendix:fine-tune parameters}
\end{table}

\section{Computing Infrastructure}
All experiments were conducted on a single NVIDIA A100 GPU (80GB). Total compute time: approximately 30 GPU-hours. Translation inference with Bloomz and Llama, as well as fine-tuning Llama were performed on this setup.
\section{Use of AI Assistants}
AI assistants were used for editing prose and debugging code. All scientific claims, experimental design, and analysis were conducted by the authors. 
\begin{figure*}[t]
\section{Fine-tuned and Base Llama BLEU and COMET Heatmaps} 
\label{Appendix fine-tune heatmaps}
  \centering\small
  \begin{tabular}{m{0.49\linewidth}m{0.49\linewidth}}
    \centering\includegraphics[width=\linewidth]{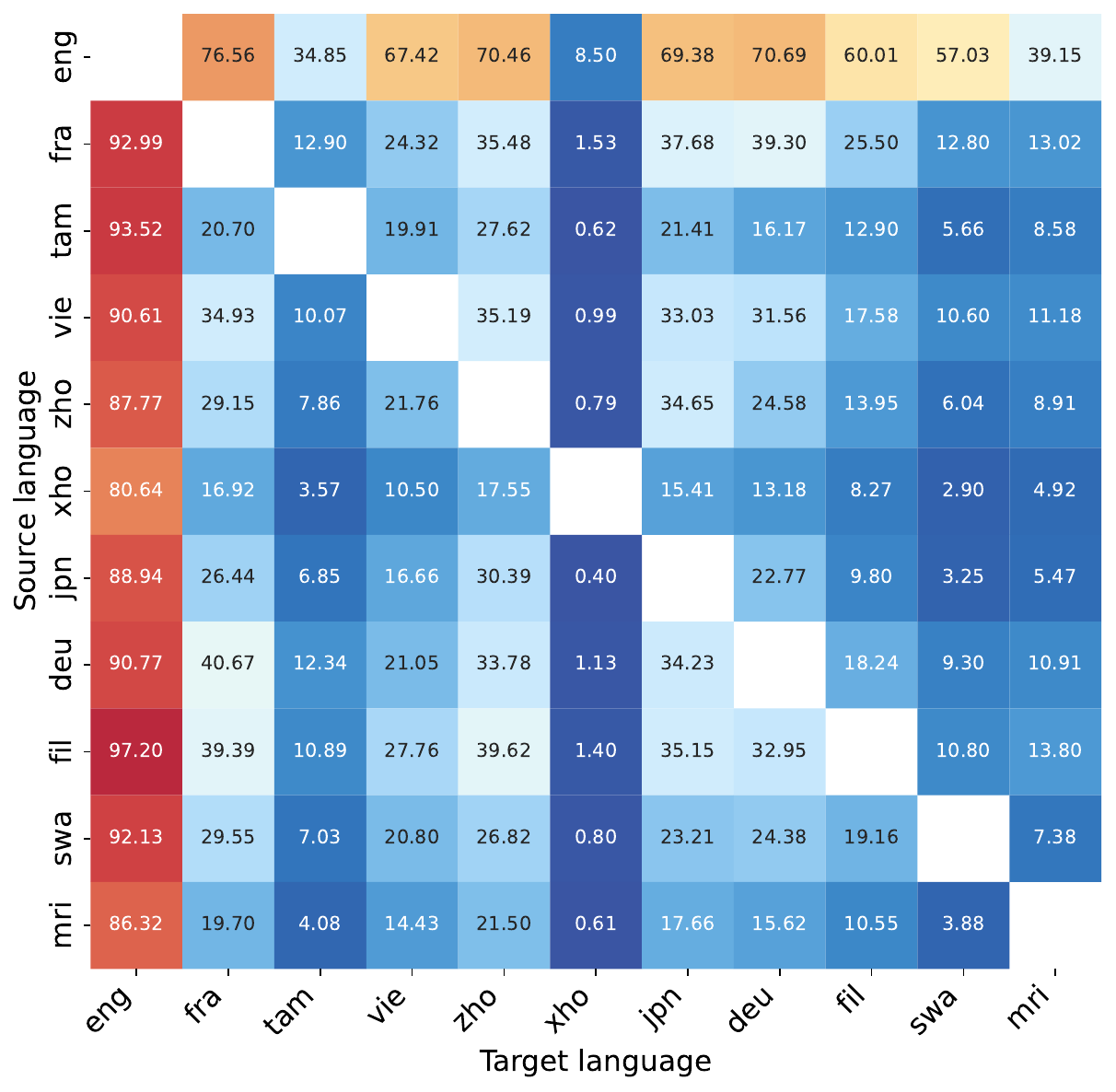}
    \centering\includegraphics[width=\linewidth]{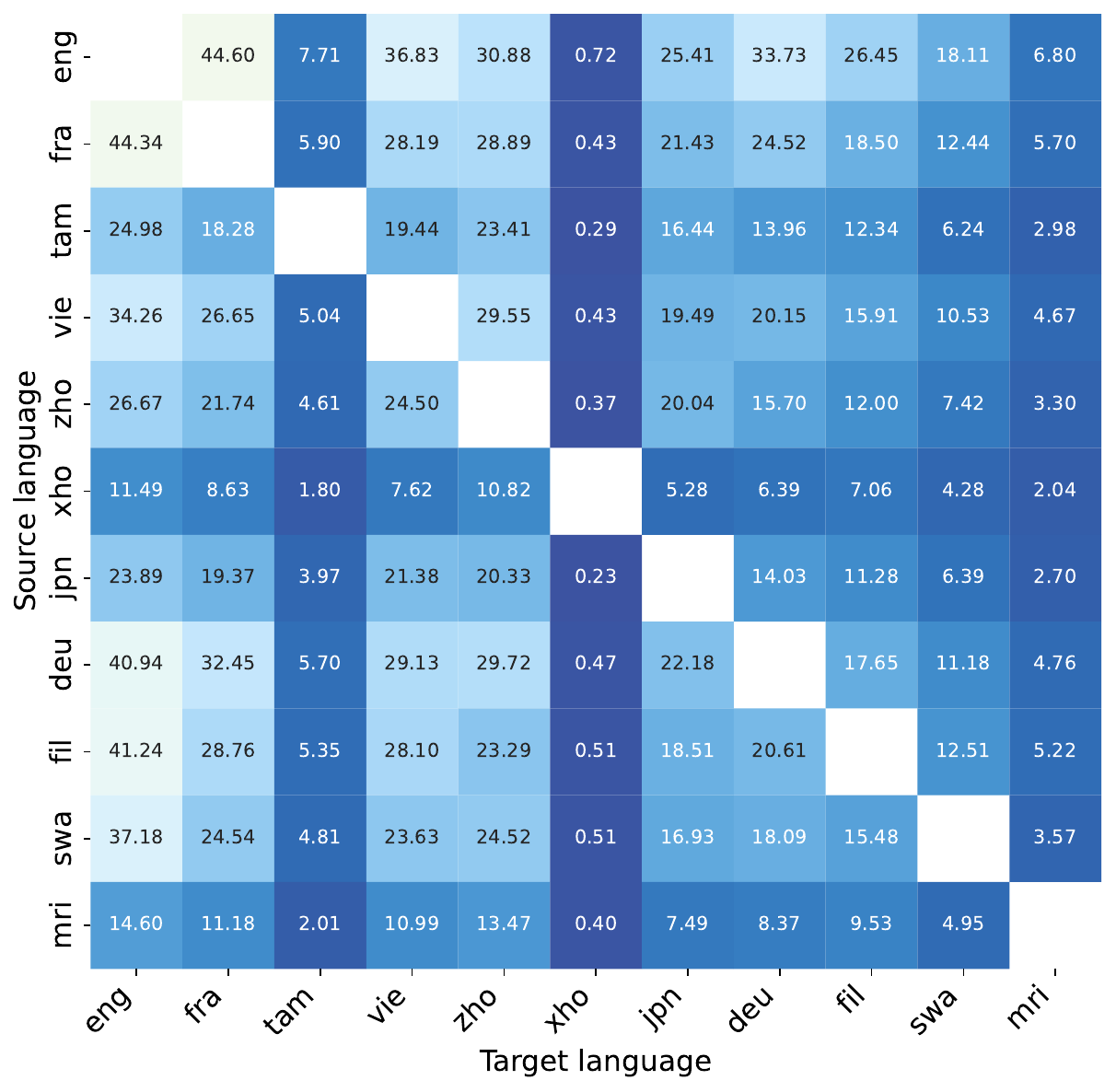} &
    \centering\includegraphics[width=\linewidth]{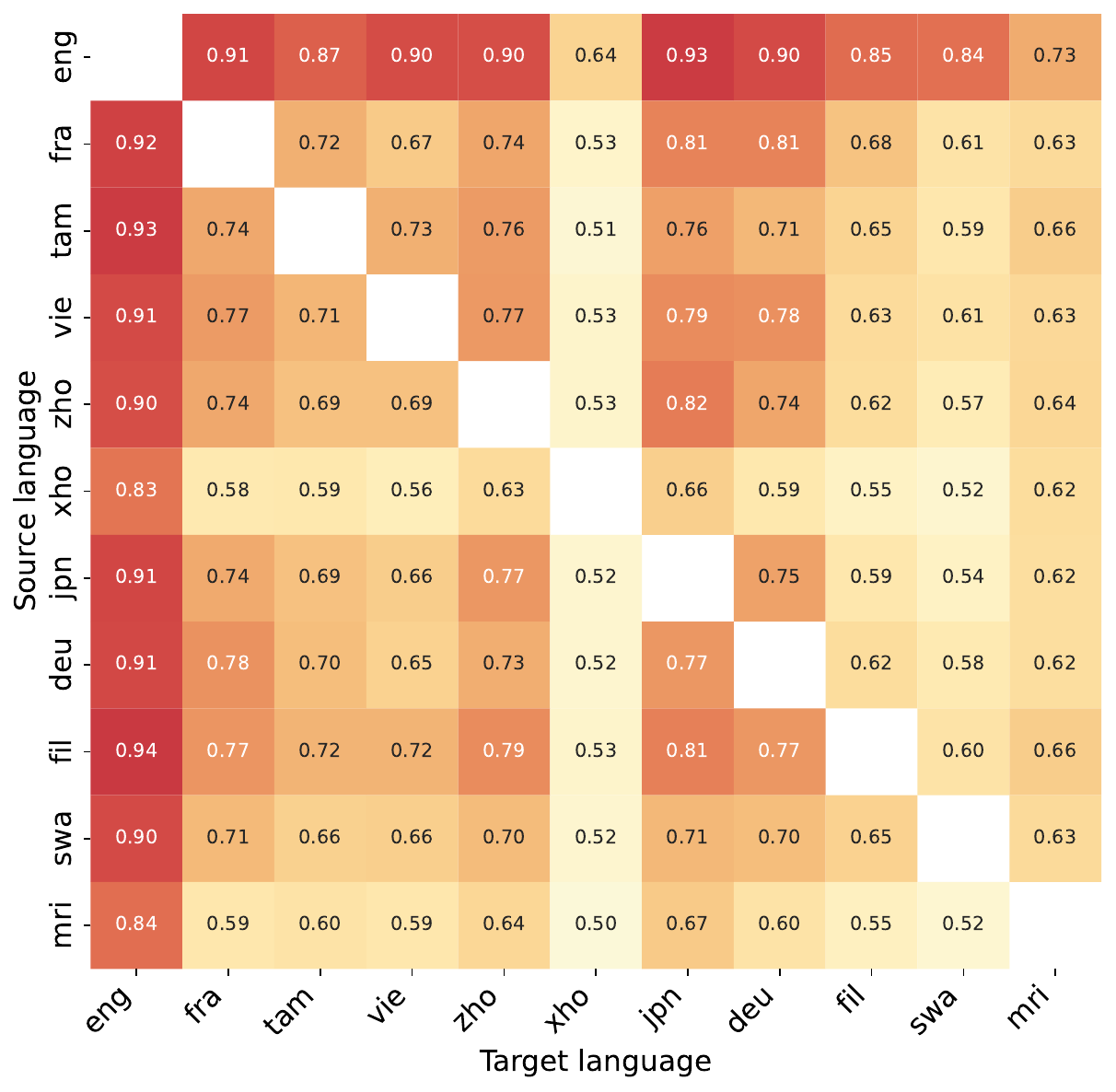}
    \centering\includegraphics[width=\linewidth]{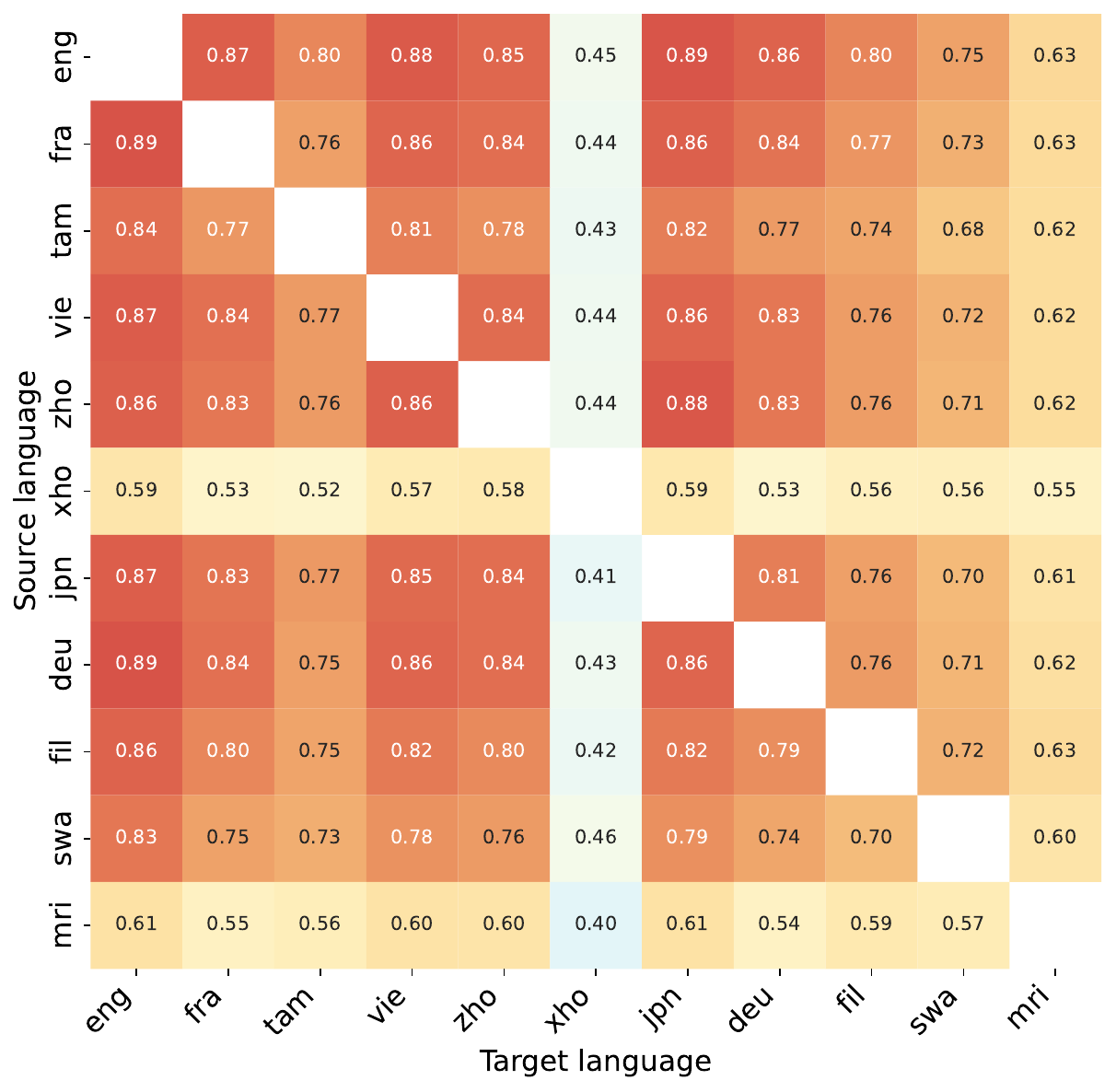}
  \end{tabular}     
  \caption{BLEU scores for the fine-tuned Llama (upper left), base Llama (lower left) and COMET scores for the fine-tuned Llama (upper right) and base Llama (lower right).}
  \label{fig:llama_ft_heatmaps_appendix}
\end{figure*}

\begin{table*}[t]
\section{Examples of Llama Back-Translations}
\label{Appendix llama backtrans}
\vspace{2ex}
\centering\small
\includegraphics[trim={3cm 3.8cm 3cm 3.8cm},clip,width=0.98\linewidth]{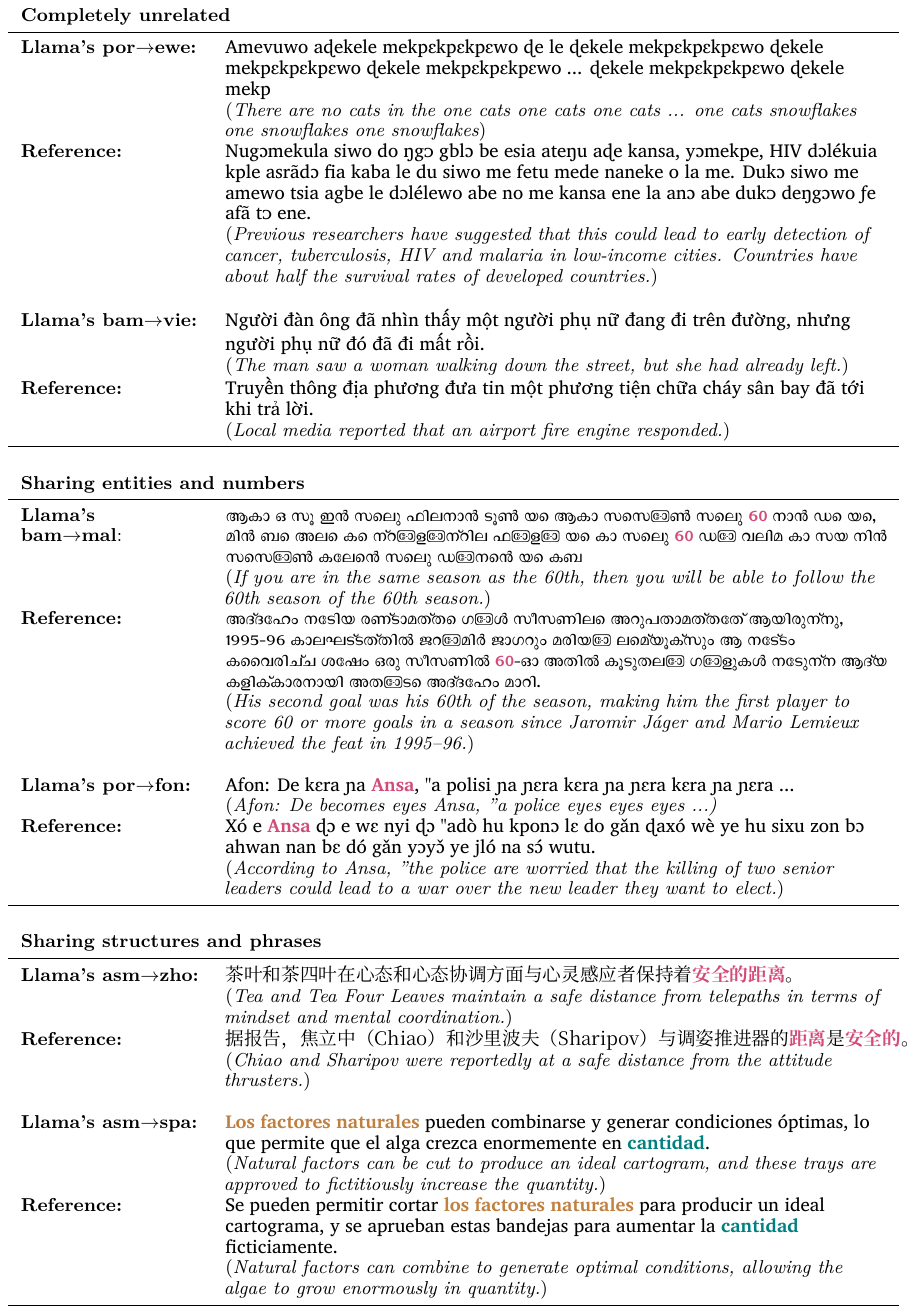}
\caption{Examples of Llama's back-translated source (\textit{with English translations from Google Translate}).}
\label{tbl: output examples}
\end{table*}

\begin{table*}[t]
\section{Examples of Entity Replaced Sentences}
\label{Appendix entity-replaced examples}
\vspace{2ex}
\centering\small
\includegraphics[trim={3cm 3.8cm 3cm 3.8cm},clip,width=0.98\linewidth]{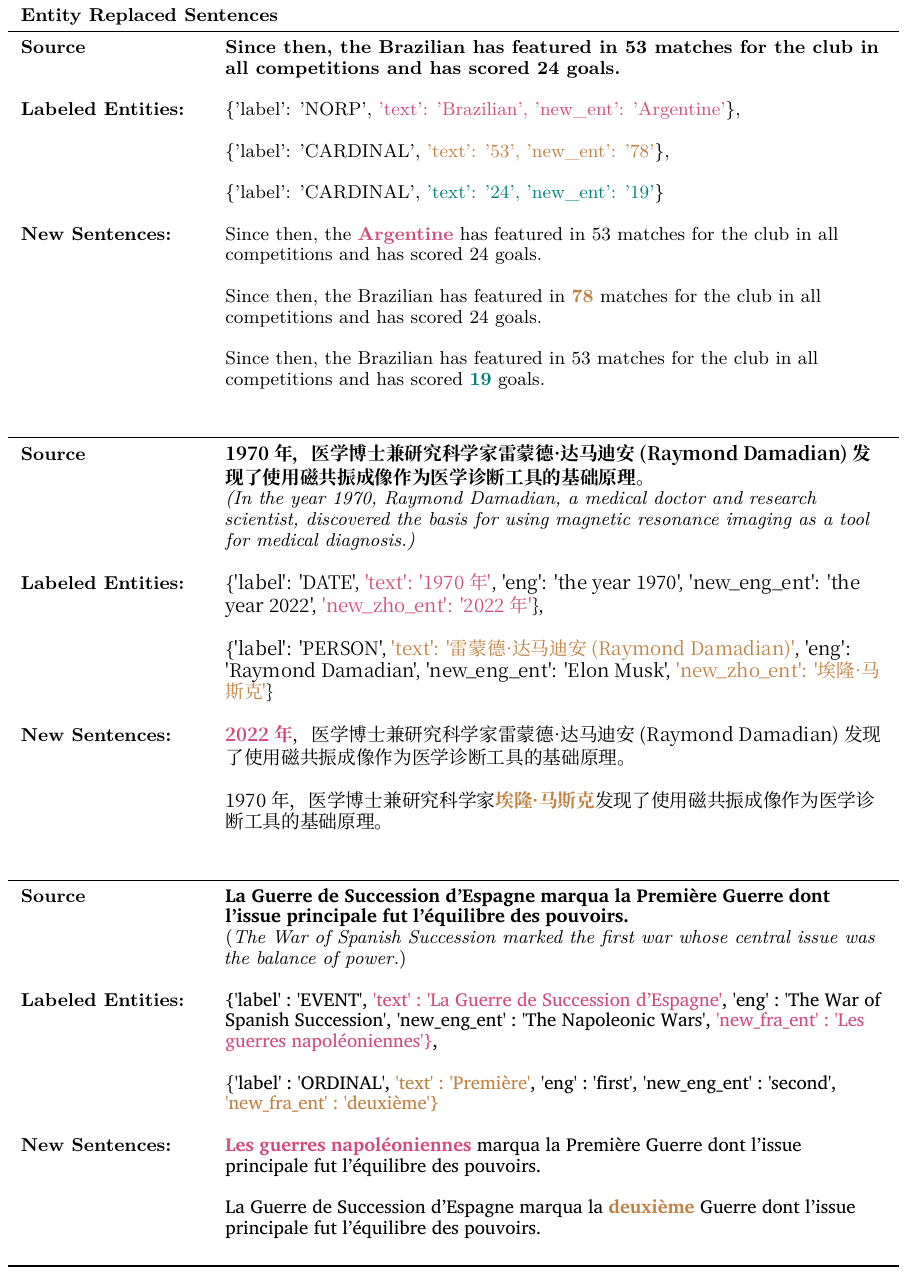}
\caption{Examples of entity replaced sentences.}
\label{tbl: entity replaced examples}
\end{table*}

\begin{table*}[t]
\section{Entity Descriptions}
\label{Appendix entity descriptions}
\centering\small
\begin{tabular}{rl}
\toprule
CARDINAL & Numerals that do not fall under another type\\
DATE & Absolute or relative dates or periods\\
EVENT & Named hurricanes, battles, wars, sports events, etc.\\
FAC & Buildings, airports, highways, bridges, etc.\\
GPE & Countries, cities, states\\
LANGUAGE & Any named language\\
LAW & Named documents made into laws.\\
LOC & Non-GPE locations, mountain ranges, bodies of water\\
MONEY & Monetary values, including unit\\
NORP & Nationalities or religious or political groups\\
ORDINAL & "first", "second", etc.\\
ORG & Companies, agencies, institutions, etc.\\
PERCENT & Percentage, including ``\%''\\
PERSON & People, including fictional\\
PRODUCT & Objects, vehicles, foods, etc. (not services)\\
QUANTITY & Measurements, as of weight or distance\\
TIME & Times smaller than a day\\
WORK\_OF\_ART & Titles of books, songs, etc.\\
MISC & Miscellaneous entities, e.g. events, nationalities, products or works of art\\
\bottomrule
\end{tabular}
\caption{Description of each Entity}
\label{tab:appendix:entity_descriptions}
\end{table*}

\begin{table*}[t]
\section{Prompts}

\vspace{2ex}
\centering
\small
\begin{minipage}{0.85\linewidth}
\begin{promptblock}
f"{{{sent}}}\nCan you translate this to tgt_lang?"
\end{promptblock}
\vspace{-2ex}
\caption{Prompt for translation with Bloomz.}

\vspace{2ex}
\begin{promptblock}
{
    "role": "system", 
    "content": "You are a helpful assistant that translates text. Your task is to translate a sentence that will be provided."
},
{
    "role": "user", 
    "content": f"Translate the following {src_name} sentence to {tgt_name}. Your response should contain only the translation and be structured like this: {{{{Your response goes here}}}}\n{sent}"
}
\end{promptblock}
\vspace{-2ex}
\caption{Prompt for translation with Llama.}

\vspace{2ex}
\begin{promptblock}
f"You are a native {lang_name} speaker.\n"
f"Task: Rephrase the following {lang_name} sentence.\n"
f"Constraints:\n"
f"- Output only one sentence.\n"
f"- Do not include explanations, notes, or formatting.\n"
f"- Do not repeat the input.\n"
f"- Do not add any extra characters or line breaks.\n\n"
f"Input:\n{src}\n\n"
f"Output:"
\end{promptblock}
\vspace{-2ex}
\caption{Prompt for paraphrasing with Llama.}

\vspace{2ex}
\label{tab:appendix:auto_term_extraction}
\begin{promptblock}
{
                "role": "user",
                "content": (
                    f"You will be given a parallel English sentence and its corresponding "
                    f"{tgt_name} sentence. The labeled named entities of the English sentence "
                    "will be provided.\n\n"

                    "Your task:\n"
                    "- For each English entity, identify the exact surface form in the "
                    f"{tgt_name} sentence that refers to the same real-world entity.\n"
                    "- DO NOT add new entities.\n"
                    "- The number of returned entities and the labels MUST match the English list.\n"
                    "- Return each entity span exactly as it appears in the target sentence.\n"
                    "- If an English entity appears multiple times, include all of them.\n"
                    "- If the English entity list is empty, return an empty list: []\n\n"

                    "Your output MUST be valid JSON ONLY, with this exact structure:\n\n"
                    "{\n"
                    f'  "entities_{tgt_lang_code}": [\n'
                    '    {"label": "ENTITY_LABEL", "text": "ENTITY_IN_TARGET_LANGUAGE", "eng": "ENTITY_IN_ENGLISH"}\n'
                    "  ]\n"
                    "}\n\n"

                    f"English Sentence:\n{src}\n\n"
                    f"English Entities:\n{ent_eng}\n\n"
                    f"{tgt_name} Sentence:\n{ref}\n\n"
                    "Output JSON:"
                )
            }
\end{promptblock}
\vspace{-2ex}
\caption{Prompt for entity alignment with Aya.}
\label{tab:appendix:entity_alignment}
\end{minipage}
\end{table*}

\begin{table*}
\vspace{2ex}
\centering
\small
\begin{minipage}{0.85\linewidth}
\begin{promptblock}
{
                "role": "user",
                "content": (
                    "You are given a list of named entities in English.\n"
                    "For each entity, generate a NEW entity of the SAME label type.\n\n"

                    "RULES:\n"
                    "- Do NOT add or remove entities.\n"
                    "- 'new_ent' must be a completely different real entity from the original.\n"
                    "- Do NOT use entities found in the original 'text' value.\n"
                    "- Synonyms, paraphrases, abbreviations, variants, or anything derived from the original are NOT allowed.\n"
                    "- The new entity must be real, valid for its label, and not share the same referent.\n"
                    "- Take the grammar structure of the entity into account if it is multiple words long.\n"
                    "- Output JSON only.\n\n"

                    
                    "OUTPUT FORMAT:\n"
                    "[{\"label\": \"LABEL\", \"text\": \"ORIGINAL_TEXT_FROM_ENTITIES_LIST\", \"new_ent\": \"NEW_ENTITY\"}]\n\n"

                    f"Entities:{ent_list}\n\n"
                    "Output JSON only:"
                )
            }

\end{promptblock}
\vspace{-2ex}
\caption{Prompt for new entity generation with Aya.}
\label{tab:appendix:new_entity_generation}

\vspace{2ex}
\begin{promptblock}
{
                "role": "user",
                "content": (
                    f"Translate the following English *ENTITY* into {tgt_name}.\n\n"

                    "RULES:\n"
                    "- Your translation MUST be grammatically correct **for the position the entity occupies** "
                    "in the target-language sentence.\n"
                    "- Imagine the translated entity being inserted into the sentence below.\n\n"

                    "RESTRICTIONS:\n"
                    "- Translate ONLY the entity.\n"
                    "- Do NOT output the whole sentence.\n"
                    "- Do NOT rewrite or paraphrase anything.\n"
                    "- No explanations.\n"
                    "- No punctuation or quotes.\n"
                    "- Output ONLY the final translated entity.\n\n"

                    f"English Sentence (Original):\n{sources[i]}\n\n"
                    f"English Sentence After Replacement:\n{eng_replaced_list[i]}\n\n"
                    f"Target-Language Reference Sentence:\n{references[i]}\n\n"
                    f"ENTITY TO TRANSLATE:\n{new_ents[i]}\n\n"

                    "Output ONLY the translated entity:"
                )
            }
\end{promptblock}
\vspace{-2ex}
\caption{Prompt for translating new entities with Aya.}
\label{tab:appendix:new_entity_translation}
\end{minipage}
\end{table*}

\end{document}